%% file: iclr2025_conference.tex
\documentclass{article} 
\usepackage{iclr2025_conference,times}
\usepackage{algorithm}
\usepackage{algpseudocode}
\usepackage{makecell}
\usepackage{hyperref}
\usepackage{url}
\usepackage{graphicx}

\input{math_commands.tex}

\title{Acceleration Multiple Heads Decoding for LLM via Dynamic Tree Attention}

\author{Zhendong Zhang \\
\texttt{zhd.zhang.ai@gmail.com} \\
}

\iclrfinalcopy 
\begin{document}
\maketitle

\begin{abstract}
   Multiple heads decoding accelerates the inference of Large Language Models (LLMs) by predicting next several tokens simultaneously.
   It generates and verifies multiple candidate sequences in parallel via tree attention with a fixed structure. 
   In this paper, we replace the fixed tree attention with dynamic tree attention on multiple head decoding, specifically in the context of MEDUSA. 
   We propose a simple and low complexity strategy to generate candidates and construct the dynamic tree structure. 
   Preliminary experiments show that the proposed method improves the decoding efficiency of multiple head decoding for LLMs 
   while maintaining the generation quality. This result demonstrates the potential for improvement of multiple head decoding in candidate generation.
\end{abstract}

\section{Introduction}
The scale of Large Language Models (LLMs) has been growing rapidly in recent years 
\citep{Radford2019LanguageMA,Brown2020LanguageMA,Achiam2023GPT4TR}. However,
this growth leads to an increase in inference latency. From a system perspective, 
the main latency bottleneck of LLM inference is memory bandwidth rather than
arithmetic computations \citep{Shazeer2019FastTD}. This bottleneck is inherent to
the sequential nature of auto-regressive decoding, which generates only a single token at a time, 
underutilizes the arithmetic computation potential of modern accelerators \citep{Cai2024MedusaSL}.

Researchers have explored generating multiple tokens simultaneously which follows a guess-verify approach. 
Depending on the methods used for initial token guessing and subsequent verification, recent techniques can be classified into three categories: 
speculative decoding, Jacobi decoding and multiple heads decoding. 
Speculative decoding uses a smaller draft model to generate a token sequence, 
which is subsequently verified by the original model \citep{Leviathan2022FastIF,Chen2023AcceleratingLL}.
Jacobi decoding typically initiates a new sequence with $[PAD]$ tokens, then iteratively verifies and updates the sequence by solving Jacobi equations 
until a fixed point is reached \citep{Song2020AcceleratingFC,Santilli2023AcceleratingTI}. 
Multiple heads decoding predicts multiple next tokens by extra output heads. 
It then constructs multiple candidate sequences by combining these heads, and verifies them in parallel by the original model 
\citep{Stern2018BlockwisePD,Cai2024MedusaSL}.

This paper focus on multiple heads decoding, particularly MEDUSA proposed in \citep{Cai2024MedusaSL}.
Given the original model's last hidden states $\mathbf{h}_t$ at last input position $t$, MEDUSA introduces $K$ additional decoding heads
to $h_t$. The $k$-th head is designed to predict the token in the $(t + k + 1)$-th position, 
while the original head predicts the $(t + 1)$-th position. Denote $\mathbf{p}^{(k)}$ as the predicted vocabulary distribution of $k$-th head.
The top predictions from $\mathbf{p}^{(k)}$ are used to generate candidate sequences, 
with each candidate being a combination of the top predictions from different heads.
MEDUSA uses a fixed set of combining patterns. By merging their common parts, these patterns are represented as a tree.
This tree structure is constructed by estimation of the accuracy via a calibration dataset. After generating candidates using the fixed tree
structure, MEDUSA verifies them in parallel through tree attention \citep{Miao2023SpecInferAG,Cai2024MedusaSL},
which involves incorporating the tree structure into the attention mask.

Although the fixed tree structure captures certain inherent biases of MEDUSA heads, it may not fully account for context-dependent variations.
We believe that a dynamic tree structure can handle context dependency better and improve the decoding efficiency of LLMs. 
In this paper, we propose a simple and efficient strategy to dynamically construct the tree structure: 
selecting top-$n$ candidates from all possible combinations (we will show how to efficiently do this). 
Experiments demonstrate that dynamic tree improves the decoding efficiency in terms of tokens per inference.
Our code is available at \url{https://github.com/zzd1992/MEDUSA-Plus}.

\section{Methodology}

\begin{algorithm}[t]
   \caption{Candidate Generation}
   \label{alg}
   \begin{algorithmic}
   
   \Require Top-$m$ probability of $K$ MEDUSA heads $\mathbf{P} \in \mathbb{R}^{K \times m}$ and number of candidates $n$
   \Ensure candidate set $S$ 
   
   \State Initialize candidate set $S \in \emptyset$
   \For{$i = 1$ to $m$}
      \State Add $(\mathbf{P}[1, i], i, 1)$ to $S$
   \EndFor
   
   \For{$k = 2$ to $K$}
      \State Initialize a priority queue $Q$ with maximum size $n$
      \For{$(p, idx, depth) \in S$}
         \State Push $(p, idx, depth)$ to $Q$
         \If{$depth=k-1$}
         \For{$i = 1$ to $m$}
            \State Push $(p \cdot \mathbf{P}[k, i], idx \times m + i, k)$ to $Q$
         \EndFor
         \EndIf
      \EndFor
      \State $S \leftarrow set($Q$)$
   \EndFor
   
   \State Return $S$
   
\end{algorithmic}
\end{algorithm}

The proposed method first dynamically generates candidates, then prepares the buffers of dynamic tree attention based on those candidates.
Ideally, candidates should be sampled according to their joint distribution. However, it is not directly accessible. 
As an alternative, we approximate the joint distribution using the Cartesian product of marginal distributions, which is provided by MEDUSA heads. Let $p_{i}^{(k)}$ 
be the $i$-th top prediction of $k$-th MEDUSA head. Then the probability of sequence $(i_1, i_2, \dots i_k)$ is
\begin{equation}
   P(i_1, i_2, \dots i_k) = \prod_{j=1}^{k} p_{i_j}^{(j)}
\end{equation}
By emulating the Cartesian Product of marginal distributions, we generate all possible candidates. 
We only consider the top-$m$ predictions of each marginal distribution, i.e. there are $\sum_{k=1}^K m^k$ possible candidates. 
Then we select top-$n$ candidates with the highest probability. This can be done efficiently by a priority queue, 
as shown in algorithm \ref{alg}. The computational complexity is $O(K nm \log n)$. 
Following \citep{Cai2024MedusaSL}, we set $K=4, n=64, m=32$. 
Thus, the actual complexity for candidates generating is quite small. 
The selected candidates form the structure of a tree due to the Cartesian Product of marginal distributions. 
\begin{equation}
   P(i_1, i_2, \dots i_k) = P(i_1, i_2, \dots i_{k-1}) P(i_k) \le P(i_1, i_2, \dots i_{k-1})
\end{equation}
If candidate $(i_1, i_2, \dots i_k)$ is selected, then it's parent $(i_1, i_2, \dots i_{k-1})$ is also selected. 
Once the candidates are generated, we prepare the buffers of dynamic tree attention, specifically the position embedding and attention mask. 
This is achieved by emulating the generated candidates with a computational complexity of $O(K n)$.
Thus, the overall computational complexity is quite small.

\section{Experiments}
We evaluate the proposed method in terms of tokens per inference (i.e. speed up) and generation quality. 
The evaluation is carried out using MT-Bench \citep{NEURIPS2023_91f18a12}, a multi-turn, conversational-format benchmark.
We use the same model of MEDUSA-1 and MEDUSA-2 which are trained with fixed backbone and trainable backbone respectively.
Currently, our evaluation is limited to Vicuna-7B model \citep{vicuna2023}. 
Generation quality is measured using the single judgment method on MT-Bench. The results are presented in Table \ref{tab}. 
The proposed method improves the decoding efficiency of MEDUSA-1 and MEDUSA-2 while maintaining the generation quality.
We provide a visualization of tree attention mask in Figure \ref{mask}. 
The dynamic tree structure shares common parts with the fixed tree structure, but can adapt to context dependencies. 
In terms of tokens per second, our method is approximately 10\% slower than MEDUSA. 
However, our implementation is not yet optimized. 
We are confident that the overhead associated with dynamic tree construction can be substantially reduced through further engineering efforts.

\begin{table}[t]
   \caption{MT-Bench results with Vicuna-7B model}
   \label{tab}
   \centering
   \begin{tabular}{ccc}
      \hline
      \hline
      Method & Speed up & Generation Quality \\
      \hline
      MEDUSA-1                & 2.50 & 5.21 \\
      MEDUSA-1 Dynamic        & 2.66 & 5.17 \\
      \hline
      MEDUSA-2                & 3.32 & 5.24 \\
      MEDUSA-2 Dynamic        & 3.51 & 5.21 \\
      \hline
   \end{tabular}
\end{table}

\begin{tabular}[t]{cccc}
   \label{mask}
   \thead{Prompt} & 
   &  
   \thead{Compose an engaging travel blog \\ post about a recent trip to Hawaii, \\ highlighting cultural experiences \\ and must-see attractions.} & 
   \thead{Describe a vivid and unique character, \\ using strong imagery and creative \\ language. Please answer in \\ fewer than two paragraphs.} \\ 
   \thead{MEDUSA-1} & 
   \thead{\includegraphics[width=0.27\columnwidth]{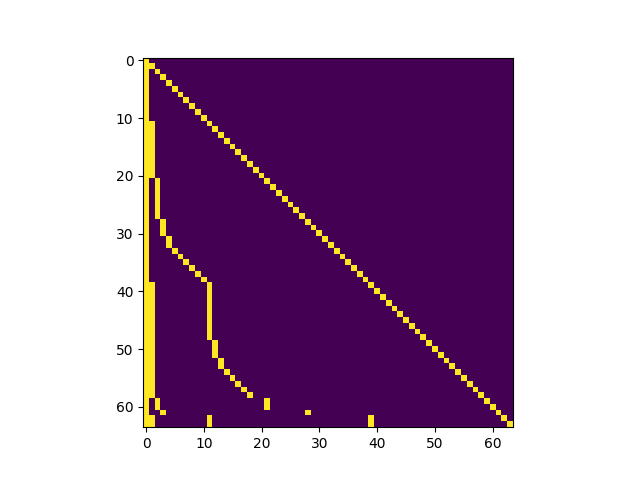}} &
   \thead{\includegraphics[width=0.27\columnwidth]{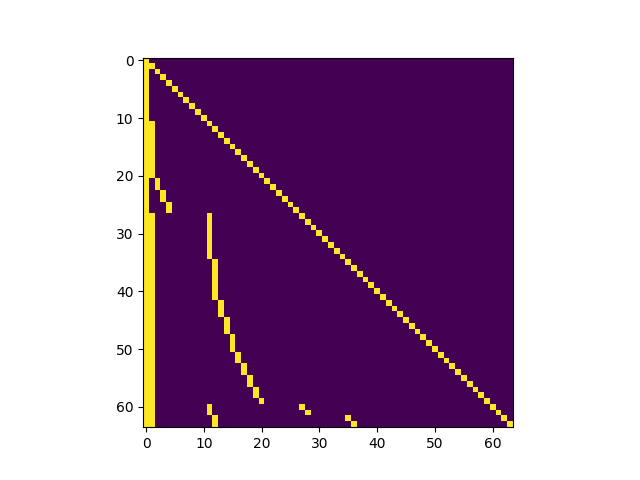}} &
   \thead{\includegraphics[width=0.27\columnwidth]{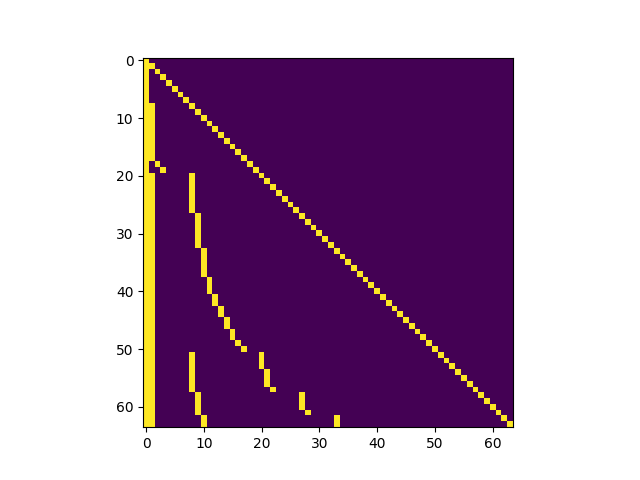}} \\
   \thead{MEDUSA-2} & 
   \thead{\includegraphics[width=0.27\columnwidth]{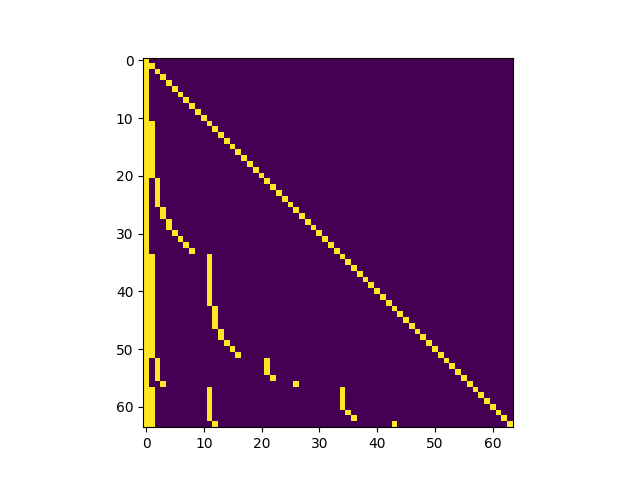}} &
   \thead{\includegraphics[width=0.27\columnwidth]{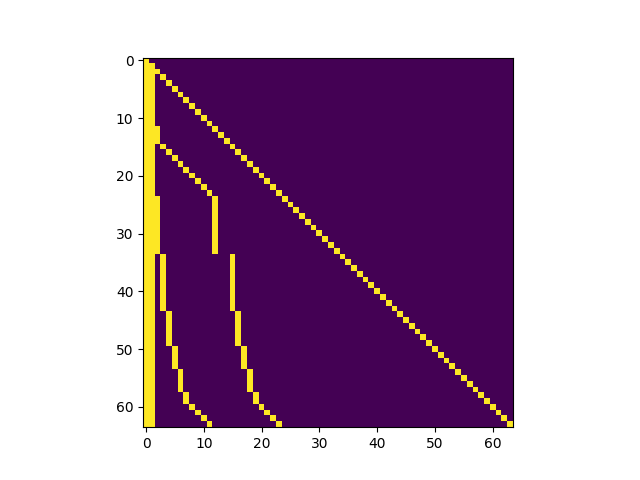}} &
   \thead{\includegraphics[width=0.27\columnwidth]{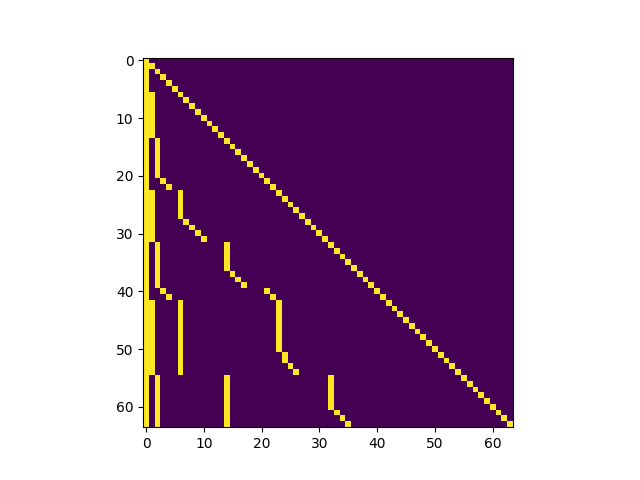}} \\
   & Fixed & Dynamic & Dynamic \\
\end{tabular}

\section{Discussion}
In this paper, we replace the fixed tree attention with dynamic tree attention on multiple head decoding, specifically in the context of MEDUSA. 
We propose a simple and low complexity strategy to generate candidates and construct the dynamic tree structure.
Preliminary experiments show that the proposed method improves the decoding efficiency of multiple head decoding for LLMs
while maintaining the generation quality.
This result demonstrates the potential for improvement of multiple head decoding in candidate generation.
For future work, we plan to improve the proposed method in the following ways:
\begin{itemize}
   \item \textbf{Optimize the overhead}: optimize candidate generation process to make our method more competitive in terms of wall time.
   \item \textbf{Improve joint distribution approximation}: currently, the joint distribution is approximated by the Cartesian product of marginal distributions. 
   We will explore better strategies to approximate it.
\end{itemize}

\bibliography{iclr2025_conference}
\bibliographystyle{iclr2025_conference}

\appendix

\end{document}

%% file: math_commands.tex
\usepackage{amsmath,amsfonts,bm}

\def\eqref#1{equation~\ref{#1}}

\def\1{\bm{1}}

\DeclareMathAlphabet{\mathsfit}{\encodingdefault}{\sfdefault}{m}{sl}
\SetMathAlphabet{\mathsfit}{bold}{\encodingdefault}{\sfdefault}{bx}{n}













%% file: iclr2025_conference.bbl
\begin{thebibliography}{13}
\providecommand{\natexlab}[1]{#1}
\providecommand{\url}[1]{\texttt{#1}}
\expandafter\ifx\csname urlstyle\endcsname\relax
  \providecommand{\doi}[1]{doi: #1}\else
  \providecommand{\doi}{doi: \begingroup \urlstyle{rm}\Url}\fi

\bibitem[Achiam et~al.(2023)Achiam, Adler, Agarwal, Ahmad, Akkaya, Aleman,
  Almeida, Altenschmidt, Altman, Anadkat, Avila, Babuschkin, Balaji, Balcom,
  Baltescu, ing Bao, Bavarian, Belgum, Bello, Berdine, Bernadett-Shapiro,
  Berner, Bogdonoff, Boiko, laine Boyd, Brakman, Brockman, Brooks, Brundage,
  Button, Cai, Campbell, Cann, Carey, Carlson, Carmichael, Chan, Chang,
  Chantzis, Chen, Chen, Chen, Chen, Chen, Chess, Cho, Chu, Chung, Cummings,
  Currier, Dai, Decareaux, Degry, Deutsch, Deville, Dhar, Dohan, Dowling,
  Dunning, Ecoffet, Eleti, Eloundou, Farhi, Fedus, Felix, Fishman, Forte,
  abella Fulford, Gao, Georges, Gibson, Goel, Gogineni, Goh, Gontijo-Lopes,
  Gordon, Grafstein, Gray, Greene, Gross, Gu, Guo, Hallacy, Han, Harris, He,
  Heaton, hannes Heidecke, Hesse, Hickey, Hickey, Hoeschele, Houghton, Hsu, Hu,
  Hu, Huizinga, Jain, Jain, Jang, Jiang, Jiang, Jin, Jin, Jomoto, Jonn, Jun,
  Kaftan, Kaiser, Kamali, Kanitscheider, Keskar, Khan, Kilpatrick, Kim, Kim,
  Kim, Kirchner, Kiros, Knight, Kokotajlo, Kondraciuk, Kondrich,
  Konstantinidis, Kosic, Krueger, Kuo, Lampe, Lan, Lee, Leike, Leung, Levy, Li,
  Lim, Lin, Lin, teusz Litwin, Lopez, Lowe, Lue, Makanju, Malfacini, Manning,
  Markov, Markovski, Martin, Mayer, Mayne, McGrew, McKinney, McLeavey,
  McMillan, McNeil, Medina, Mehta, Menick, Metz, Mishchenko, Mishkin, Monaco,
  Morikawa, Mossing, Mu, Murati, Murk, M'ely, Nair, Nakano, Nayak, Neelakantan,
  Ngo, Noh, Long, O'Keefe, Pachocki, Paino, Palermo, Pantuliano, Parascandolo,
  Parish, Parparita, Passos, Pavlov, Peng, Perelman, de~Avila Belbute~Peres,
  Petrov, de~Oliveira~Pinto, Pokorny, Pokrass, Pong, Powell, Power, Power,
  Proehl, Puri, Radford, Rae, Ramesh, Raymond, Real, Rimbach, Ross, Rotsted,
  Roussez, Ryder, Saltarelli, Sanders, Santurkar, Sastry, Schmidt, Schnurr,
  Schulman, Selsam, Sheppard, Sherbakov, Shieh, Shoker, Shyam, Sidor, Sigler,
  Simens, Sitkin, Slama, Sohl, Sokolowsky, Song, Staudacher, Such, Summers,
  Sutskever, Tang, Tezak, Thompson, Tillet, Tootoonchian, Tseng, Tuggle,
  Turley, Tworek, Uribe, Vallone, Vijayvergiya, Voss, Wainwright, Wang, Wang,
  Wang, Ward, Wei, Weinmann, Welihinda, Welinder, Weng, Weng, Wiethoff,
  Willner, Winter, Wolrich, Wong, Workman, Wu, Wu, Wu, Xiao, Xu, Yoo, Yu, ing
  Yuan, Zaremba, Zellers, Zhang, Zhang, Zhao, Zheng, Zhuang, Zhuk, and
  Zoph]{Achiam2023GPT4TR}
OpenAI~Josh Achiam, Steven Adler, Sandhini Agarwal, Lama Ahmad, Ilge Akkaya,
  Florencia~Leoni Aleman, Diogo Almeida, Janko Altenschmidt, Sam Altman,
  Shyamal Anadkat, Red Avila, Igor Babuschkin, Suchir Balaji, Valerie Balcom,
  Paul Baltescu, Haim ing Bao, Mo~Bavarian, Jeff Belgum, Irwan Bello, Jake
  Berdine, Gabriel Bernadett-Shapiro, Christopher Berner, Lenny Bogdonoff, Oleg
  Boiko, Made laine Boyd, Anna-Luisa Brakman, Greg Brockman, Tim Brooks, Miles
  Brundage, Kevin Button, Trevor Cai, Rosie Campbell, Andrew Cann, Brittany
  Carey, Chelsea Carlson, Rory Carmichael, Brooke Chan, Che Chang, Fotis
  Chantzis, Derek Chen, Sully Chen, Ruby Chen, Jason Chen, Mark Chen, Benjamin
  Chess, Chester Cho, Casey Chu, Hyung~Won Chung, Dave Cummings, Jeremiah
  Currier, Yunxing Dai, Cory Decareaux, Thomas Degry, Noah Deutsch, Damien
  Deville, Arka Dhar, David Dohan, Steve Dowling, Sheila Dunning, Adrien
  Ecoffet, Atty Eleti, Tyna Eloundou, David Farhi, Liam Fedus, Niko Felix,
  Sim'on~Posada Fishman, Juston Forte, Is~abella Fulford, Leo Gao, Elie
  Georges, Christian Gibson, Vik Goel, Tarun Gogineni, Gabriel Goh, Raphael
  Gontijo-Lopes, Jonathan Gordon, Morgan Grafstein, Scott Gray, Ryan Greene,
  Joshua Gross, Shixiang~Shane Gu, Yufei Guo, Chris Hallacy, Jesse Han, Jeff
  Harris, Yuchen He, Mike Heaton, Jo~hannes Heidecke, Chris Hesse, Alan Hickey,
  Wade Hickey, Peter Hoeschele, Brandon Houghton, Kenny Hsu, Shengli Hu, Xin
  Hu, Joost Huizinga, Shantanu Jain, Shawn Jain, Joanne Jang, Angela Jiang,
  Roger Jiang, Haozhun Jin, Denny Jin, Shino Jomoto, Billie Jonn, Heewoo Jun,
  Tomer Kaftan, Lukasz Kaiser, Ali Kamali, Ingmar Kanitscheider, Nitish~Shirish
  Keskar, Tabarak Khan, Logan Kilpatrick, Jong~Wook Kim, Christina Kim, Yongjik
  Kim, Hendrik Kirchner, Jamie~Ryan Kiros, Matthew Knight, Daniel Kokotajlo,
  Lukasz Kondraciuk, Andrew Kondrich, Aris Konstantinidis, Kyle Kosic, Gretchen
  Krueger, Vishal Kuo, Michael Lampe, Ikai Lan, Teddy Lee, Jan Leike, Jade
  Leung, Daniel Levy, Chak~Ming Li, Rachel Lim, Molly Lin, Stephanie Lin,
  Ma~teusz Litwin, Theresa Lopez, Ryan Lowe, Patricia Lue, Anna Makanju, Kim
  Malfacini, Sam Manning, Todor Markov, Yaniv Markovski, Bianca Martin, Katie
  Mayer, Andrew Mayne, Bob McGrew, Scott~Mayer McKinney, Christine McLeavey,
  Paul McMillan, Jake McNeil, David Medina, Aalok Mehta, Jacob Menick, Luke
  Metz, Andrey Mishchenko, Pamela Mishkin, Vinnie Monaco, Evan Morikawa,
  Daniel~P. Mossing, Tong Mu, Mira Murati, Oleg Murk, David M'ely, Ashvin Nair,
  Reiichiro Nakano, Rajeev Nayak, Arvind Neelakantan, Richard Ngo, Hyeonwoo
  Noh, Ouyang Long, Cullen O'Keefe, Jakub~W. Pachocki, Alex Paino, Joe Palermo,
  Ashley Pantuliano, Giambattista Parascandolo, Joel Parish, Emy Parparita,
  Alexandre Passos, Mikhail Pavlov, Andrew Peng, Adam Perelman, Filipe de~Avila
  Belbute~Peres, Michael Petrov, Henrique~Pond{\'e} de~Oliveira~Pinto, Michael
  Pokorny, Michelle Pokrass, Vitchyr~H. Pong, Tolly Powell, Alethea Power,
  Boris Power, Elizabeth Proehl, Raul Puri, Alec Radford, Jack~W. Rae, Aditya
  Ramesh, Cameron Raymond, Francis Real, Kendra Rimbach, Carl Ross, Bob
  Rotsted, Henri Roussez, Nick Ryder, Mario~D. Saltarelli, Ted Sanders, Shibani
  Santurkar, Girish Sastry, Heather Schmidt, David Schnurr, John Schulman,
  Daniel Selsam, Kyla Sheppard, Toki Sherbakov, Jessica Shieh, Sarah Shoker,
  Pranav Shyam, Szymon Sidor, Eric Sigler, Maddie Simens, Jordan Sitkin,
  Katarina Slama, Ian Sohl, Benjamin Sokolowsky, Yang Song, Natalie Staudacher,
  Felipe~Petroski Such, Natalie Summers, Ilya Sutskever, Jie Tang, Nikolas~A.
  Tezak, Madeleine Thompson, Phil Tillet, Amin Tootoonchian, Elizabeth Tseng,
  Preston Tuggle, Nick Turley, Jerry Tworek, Juan Felipe~Cer'on Uribe, Andrea
  Vallone, Arun Vijayvergiya, Chelsea Voss, Carroll~L. Wainwright, Justin~Jay
  Wang, Alvin Wang, Ben Wang, Jonathan Ward, Jason Wei, CJ~Weinmann, Akila
  Welihinda, Peter Welinder, Jiayi Weng, Lilian Weng, Matt Wiethoff, Dave
  Willner, Clemens Winter, Samuel Wolrich, Hannah Wong, Lauren Workman, Sherwin
  Wu, Jeff Wu, Michael Wu, Kai Xiao, Tao Xu, Sarah Yoo, Kevin Yu, Qim ing Yuan,
  Wojciech Zaremba, Rowan Zellers, Chong Zhang, Marvin Zhang, Shengjia Zhao,
  Tianhao Zheng, Juntang Zhuang, William Zhuk, and Barret Zoph.
\newblock Gpt-4 technical report.
\newblock 2023.
\newblock URL \url{https://api.semanticscholar.org/CorpusID:257532815}.

\bibitem[Brown et~al.(2020)Brown, Mann, Ryder, Subbiah, Kaplan, Dhariwal,
  Neelakantan, Shyam, Sastry, Askell, Agarwal, Herbert-Voss, Krueger, Henighan,
  Child, Ramesh, Ziegler, Wu, Winter, Hesse, Chen, Sigler, teusz Litwin, Gray,
  Chess, Clark, Berner, McCandlish, Radford, Sutskever, and
  Amodei]{Brown2020LanguageMA}
Tom~B. Brown, Benjamin Mann, Nick Ryder, Melanie Subbiah, Jared Kaplan,
  Prafulla Dhariwal, Arvind Neelakantan, Pranav Shyam, Girish Sastry, Amanda
  Askell, Sandhini Agarwal, Ariel Herbert-Voss, Gretchen Krueger, Tom Henighan,
  Rewon Child, Aditya Ramesh, Daniel~M. Ziegler, Jeff Wu, Clemens Winter,
  Christopher Hesse, Mark Chen, Eric Sigler, Ma~teusz Litwin, Scott Gray,
  Benjamin Chess, Jack Clark, Christopher Berner, Sam McCandlish, Alec Radford,
  Ilya Sutskever, and Dario Amodei.
\newblock Language models are few-shot learners.
\newblock \emph{ArXiv}, abs/2005.14165, 2020.
\newblock URL \url{https://api.semanticscholar.org/CorpusID:218971783}.

\bibitem[Cai et~al.(2024)Cai, Li, Geng, Peng, Lee, huai Chen, and
  Dao]{Cai2024MedusaSL}
Tianle Cai, Yuhong Li, Zhengyang Geng, Hongwu Peng, Jason~D. Lee, De~huai Chen,
  and Tri Dao.
\newblock Medusa: Simple llm inference acceleration framework with multiple
  decoding heads.
\newblock \emph{ArXiv}, abs/2401.10774, 2024.
\newblock URL \url{https://api.semanticscholar.org/CorpusID:267061277}.

\bibitem[Chen et~al.(2023)Chen, Borgeaud, Irving, Lespiau, Sifre, and
  Jumper]{Chen2023AcceleratingLL}
Charlie Chen, Sebastian Borgeaud, Geoffrey Irving, Jean-Baptiste Lespiau,
  L.~Sifre, and John~M. Jumper.
\newblock Accelerating large language model decoding with speculative sampling.
\newblock \emph{ArXiv}, abs/2302.01318, 2023.
\newblock URL \url{https://api.semanticscholar.org/CorpusID:256503945}.

\bibitem[Chiang et~al.(2023)Chiang, Li, Lin, Sheng, Wu, Zhang, Zheng, Zhuang,
  Zhuang, Gonzalez, Stoica, and Xing]{vicuna2023}
Wei-Lin Chiang, Zhuohan Li, Zi~Lin, Ying Sheng, Zhanghao Wu, Hao Zhang, Lianmin
  Zheng, Siyuan Zhuang, Yonghao Zhuang, Joseph~E. Gonzalez, Ion Stoica, and
  Eric~P. Xing.
\newblock Vicuna: An open-source chatbot impressing gpt-4 with 90\%* chatgpt
  quality, March 2023.
\newblock URL \url{https://lmsys.org/blog/2023-03-30-vicuna/}.

\bibitem[Leviathan et~al.(2022)Leviathan, Kalman, and
  Matias]{Leviathan2022FastIF}
Yaniv Leviathan, Matan Kalman, and Yossi Matias.
\newblock Fast inference from transformers via speculative decoding.
\newblock In \emph{International Conference on Machine Learning}, 2022.
\newblock URL \url{https://api.semanticscholar.org/CorpusID:254096365}.

\bibitem[Miao et~al.(2023)Miao, Oliaro, Zhang, Cheng, Wang, Wong, Chen, Arfeen,
  Abhyankar, and Jia]{Miao2023SpecInferAG}
Xupeng Miao, Gabriele Oliaro, Zhihao Zhang, Xinhao Cheng, Zeyu Wang, Rae
  Ying~Yee Wong, Zhuoming Chen, Daiyaan Arfeen, Reyna Abhyankar, and Zhihao
  Jia.
\newblock Specinfer: Accelerating generative llm serving with speculative
  inference and token tree verification.
\newblock \emph{ArXiv}, abs/2305.09781, 2023.
\newblock URL \url{https://api.semanticscholar.org/CorpusID:258740799}.

\bibitem[Radford et~al.(2019)Radford, Wu, Child, Luan, Amodei, and
  Sutskever]{Radford2019LanguageMA}
Alec Radford, Jeff Wu, Rewon Child, David Luan, Dario Amodei, and Ilya
  Sutskever.
\newblock Language models are unsupervised multitask learners.
\newblock 2019.
\newblock URL \url{https://api.semanticscholar.org/CorpusID:160025533}.

\bibitem[Santilli et~al.(2023)Santilli, Severino, Postolache, Maiorca, Mancusi,
  Marin, and Rodol{\`a}]{Santilli2023AcceleratingTI}
Andrea Santilli, Silvio Severino, Emilian Postolache, Valentino Maiorca,
  Michele Mancusi, Riccardo Marin, and Emanuele Rodol{\`a}.
\newblock Accelerating transformer inference for translation via parallel
  decoding.
\newblock In \emph{Annual Meeting of the Association for Computational
  Linguistics}, 2023.
\newblock URL \url{https://api.semanticscholar.org/CorpusID:258741236}.

\bibitem[Shazeer(2019)]{Shazeer2019FastTD}
Noam~M. Shazeer.
\newblock Fast transformer decoding: One write-head is all you need.
\newblock \emph{ArXiv}, abs/1911.02150, 2019.
\newblock URL \url{https://api.semanticscholar.org/CorpusID:207880429}.

\bibitem[Song et~al.(2020)Song, Meng, Liao, and Ermon]{Song2020AcceleratingFC}
Yang Song, Chenlin Meng, Renjie Liao, and Stefano Ermon.
\newblock Accelerating feedforward computation via parallel nonlinear equation
  solving.
\newblock In \emph{International Conference on Machine Learning}, 2020.
\newblock URL \url{https://api.semanticscholar.org/CorpusID:235422735}.

\bibitem[Stern et~al.(2018)Stern, Shazeer, and Uszkoreit]{Stern2018BlockwisePD}
Mitchell Stern, Noam~M. Shazeer, and Jakob Uszkoreit.
\newblock Blockwise parallel decoding for deep autoregressive models.
\newblock In \emph{Neural Information Processing Systems}, 2018.
\newblock URL \url{https://api.semanticscholar.org/CorpusID:53208380}.

\bibitem[Zheng et~al.(2023)Zheng, Chiang, Sheng, Zhuang, Wu, Zhuang, Lin, Li,
  Li, Xing, Zhang, Gonzalez, and Stoica]{NEURIPS2023_91f18a12}
Lianmin Zheng, Wei-Lin Chiang, Ying Sheng, Siyuan Zhuang, Zhanghao Wu, Yonghao
  Zhuang, Zi~Lin, Zhuohan Li, Dacheng Li, Eric Xing, Hao Zhang, Joseph~E
  Gonzalez, and Ion Stoica.
\newblock Judging llm-as-a-judge with mt-bench and chatbot arena.
\newblock In A.~Oh, T.~Naumann, A.~Globerson, K.~Saenko, M.~Hardt, and
  S.~Levine (eds.), \emph{Advances in Neural Information Processing Systems},
  volume~36, pp.\  46595--46623. Curran Associates, Inc., 2023.
\newblock URL
  \url{https://proceedings.neurips.cc/paper_files/paper/2023/file/91f18a1287b398d378ef22505bf41832-Paper-Datasets_and_Benchmarks.pdf}.

\end{thebibliography}
